\documentclass[11pt,a4paper]{article}
\usepackage[hyperref]{acl2019}
\usepackage{times}
\usepackage{latexsym}

\usepackage{url}

\aclfinalcopy 

\usepackage{nicefrac, amsmath, amssymb, amsthm, bbm, mathtools, algorithm, algorithmic}

\usepackage{tikz}
\usetikzlibrary{arrows, automata, positioning}

\newcommand{\Prob}{\mathbb{P}}
\newcommand{\Exp}{\mathbb{E}}

\newcommand\numberthis{\addtocounter{equation}{1}\tag{\theequation}}

\title{Towards Near-imperceptible Steganographic Text}

\author{Falcon Z.~Dai \\
  Toyota Technological Institute at Chicago \\
  Chicago, USA \\
  \texttt{dai@ttic.edu} \\\And
  Zheng Cai \\
  Department of Computer Science \\
  University of Colorado \\
  Boulder, CO, USA\\
  \texttt{jon.z.cai@colorado.edu} \\}

\date{}

\begin{document}

\maketitle

\begin{abstract}
  We show that the imperceptibility of several existing linguistic steganographic systems \cite{fang2017generating,yang2018rnn} relies on implicit assumptions on statistical behaviors of fluent text. 
We formally analyze them and empirically evaluate these assumptions. 
Furthermore, based on these observations, we propose an encoding algorithm called \texttt{patient-Huffman} with improved near-imperceptible guarantees.
\end{abstract}


\section{Introduction}
\label{sec:introduction}

In recent years, we see many exciting developments in applied machine learning and, in particular, its application in the fundamental problem of language modeling \cite{sutskever2011generating, jozefowicz2016exploring} in the field of natural language processing (NLP). However, these advancements can be exploited by computationally resourceful entities such as a surveillance state to effectively monitor its citizens' \emph{ostensibly} private communications at scale.

We are motivated to study the communication privacy problem of concealing sensitive messages in monitored channels.
In order to avoid raising suspicion in the monitoring party, we want to hide the intended message inside a fluent message, known as a \emph{stegotext}, indistinguishable from what is expected in such channels. This is a problem studied primarily in \emph{steganography} and steganography researchers have a keen interest in linguistic steganography as it presents fundamental challenges \cite{chang2014practical}; the linguistic channel carries few bits per symbol on average \cite{shannon1951prediction, brown1992estimate} making it hard to hide a message. In contrast, images and sound recordings have a high information theoretic entropy comparing to a written message making it relatively easy to embed a message in the noise floor of the channel.

This problem of hiding secret messages in plain sight might evoke spy stories of concealing messages on newspaper advertisements during Cold War. Such manual methods have been superseded by algorithmic approaches. Classic methods prior to the advance of applied machine learning in this domain typically try to produce \emph{grammatical} English with generative grammar \cite{chapman1997hiding}. However, such generation methods fall short in terms of \emph{statistical} imperceptibility \cite{meng2008linguistic}. This makes them vulnerable to automated detection. Generating \emph{fluent}\footnote{It is often referred to as ``naturalness'' in linguistic steganography literature.} text at scale is at the heart of the steganography problem, and language models (LM) studied in NLP provide a natural solution by letting us draw samples of fluent texts.

At the working heart of a LM-based stegosystem, there lies an encoding algorithm that encodes a ciphertext (a random string indistinguishable from a series of fair coin flips) into a fluent stegotext using an LM. From the communication standpoint, this encoding must be uniquely decodable, i.e. different ciphertext are encoded into different stegotexts otherwise the receiver will not be able to decode and recover the ciphertext. Instead of sampling according to the LM, an encoding algorithm effectively induces a new language model by providing a non-standard way to draw samples from the LM. Thus, from the language modeling standpoint, in order to achieve statistical imperceptibility, extra care is needed to ensure the resulting LM is close to the original LM (Sec.~\ref{sec:decompose}). Various uniquely decodable algorithms has been devised by recent pioneering works \cite{fang2017generating, yang2018rnn} leveraging recurrent neural network-based LMs, and the high-quality stegotexts generated show tremendous promise in terms of both fluency and information hiding capacity. However, these methods do not explicitly provide guarantees on imperceptibility. Instead, their imperceptibility, as we will argue, relies on \emph{implicit} assumptions on the statistical behaviors of the underlying LM, and ultimately, of fluent texts (Sec.~\ref{sec:analysis}). We will empirically evaluate these assumptions and show that they are problematic (Sec.~\ref{sec:bins}).
In response, we will propose an improved encoding algorithm \texttt{patient-Huffman} that explicitly maintains imperceptibility (Sec.~\ref{sec:algorithm}).

To see that imperceptibility crucially depends on the statistics of fluent texts, consider plausible continuations of the following two prefixes, ``I like your'' and ``It is on top.'' In the first case, there are many likely next words such as ``work'', ``style'', ``idea'', ``game'', ``book'', whereas in the latter, there are few such as ``of'', ``,'', ``and'', ``.'' with ``of'' being overwhelmingly likely.
Intuitively speaking, the distribution over next tokens in fluent texts is sometimes uniform and sometimes highly concentrated.\footnote{Under the estimates of \texttt{GPT-2-117M}, the first continuation has entropy of 11.2 bits and the latter, 0.43 bits. The most likely next tokens shown are also drawn from this model.} When it is concentrated, if we choose the next token by flipping fair coins, we will be sampling from a very different distribution and risk being detected after a few samples.
In \texttt{patient-Huffman}, we actively evaluate how different the encoding distribution and the LM distribution are, and avoid encoding at steps that can expose us.


The highlights of this work are the following:
\begin{itemize}
  \itemsep 0em
  \item We quantify statistical imperceptibility with total variation distance (TVD) between language models. We study the TVD of several encoding algorithms and point out the implicit assumption for them to be near-imperceptible.
  \item We use a state-of-the-art transformer-based, subword-level LM, \texttt{GPT-2-117M} \cite{radford2019language}, to empirically evaluate the plausibility of assumptions implicitly made by different encoding methods.
  \item We propose an encoding algorithm \texttt{patient-Huffman} with strong relative statistical imperceptibility.
\end{itemize}


%

\section{Formalism}

Suppose Alice (sender) wants to send Bob (receiver) a sensitive message (plaintext) through a channel monitored by Eve (adversary). This channel may be shared by many other communicating parties. Furthermore, Eve expects to see fluent natural language texts in this channel. Alice wants to avoid sending non-fluent texts through this channel to raise Eve's suspicion while ensuring that only Bob can read the sensitive message.

One class of solutions is to
\begin{enumerate}
  \itemsep 0em
  \item Alice \emph{encrypts} the plaintext message into a ciphertext with a key shared with Bob.\footnote{Public key encryption can also work. Alice will encrypt the plaintext with Bob's public key and Bob decrypts with his private key in that case.}
  \item Alice \emph{hides} the ciphertext, which has the statistics of random coin flips, into a fluent stegotext.
  \item Alice sends the stegotext through a channel monitored by Eve.
  \item Bob receives the stegotext and \emph{seeks} the ciphertext from it.
  \item Bob \emph{decrypts} the ciphertext with the shared key and obtain the plaintext message.
\end{enumerate}

Linguistic stegosystems concern with steps 2 (hide) and 4 (seek), i.e. encoding a random bitstring into a fluent stegotext and extracting the original bitstring from such fluent stegotexts, respectively.

A \emph{vocabulary} $\Sigma$ of size $V$ is a finite set of tokens.\footnote{Tokens can be characters, subword units or words depending on the modeling choices. We will be using subword units based on byte pair encoding in our experiments.}
An \emph{extended vocabulary} $\Sigma^*$ is the set of all finite sequences of tokens from $\Sigma$. We call its elements \emph{texts}.
A \emph{language model} $\ell$ is a measure over some extended vocabulary $\Sigma^*$. Furthermore, we assume that we have access to the conditional distribution over the next token given a prefix $\Prob[s_{t+1} | s_1 \cdots s_t ; \ell]$ and the distribution of the initial token $\Prob[s_1 ; \ell]$. An LM specified in this way allows us to draw samples easily. We can draw a sample text by drawing each $s_t$ one at a time for $t=1, 2, \cdots$ according to LM. We call the random sample text $s \coloneqq s_1 \cdots s_T \sim \ell$ an \emph{$\ell$-fluent text}.
 
Total variation distance (TVD) between two measures $p$ and $q$ over the same events denoted by $\sigma$-algebra $\mathcal{F}$, is $d(p, q) \coloneqq \sup_{E \in \mathcal{F}} |p(E) - q(E)|$ (see \ref{app:tvd} for more facts). 

A \emph{ciphertext} $b$ of length $C$ is a random variable $b \coloneqq b_1 b_2 \cdots b_C \sim \text{Bernoulli}(\nicefrac{1}{2})^C$. 
An \emph{encoding algorithm} $\mathfrak{A}_\ell$ is an injective map from ciphertexts to distributions over texts $\mathfrak{A}_\ell : \{0, 1\}^C \rightarrow \Delta(\Sigma^*)$ which may depend on the LM $\ell$. Injectivity ensures that the stegotexts are unique decodable. 

\subsection{Near-imperceptibility}
Instead of using the informal notion of imperceptibility common in steganography literature which relies on a human eavesdropper (playing Eve) judging the quality, we consider a formal statistical notion of near-imperceptibility. We say a measure over texts $p$, i.e. an LM, is \emph{$\delta$-imperceptible with respect to a language model $\ell$} if $d(p, \ell) < \delta$. This formalization is motivated by the fact that for any algorithm, it takes \emph{at least} $\Omega\left(\frac{1}{\delta^2}\right)$-many samples to tell whether the samples come from $\ell$ or $p$ with high confidence.\footnote{This is a basic result from information theory. See for example \cite{madhur}.} The smaller $d(p, \ell)$ is, the more samples are required for Eve to discover the presence of our steganographic communication regardless of her computational resource.
Therefore, we want to find encoding algorithms that are near-imperceptible with respect to the true LM of the monitored channel.

\subsection{Decomposition of TVD}
\label{sec:decompose}
Suppose the true LM of the monitored channel is $\ell^*$, and we have access to a base LM $\ell$, then running encoding algorithm $\mathfrak{A}_\ell$ induces an effective LM $ \mathfrak{A}[\ell] \coloneqq \Exp_b[\mathfrak{A}_\ell(b)]$. Consider the TVD between the effective LM and the true LM 
$$ d(\ell^*, \mathfrak{A}[\ell]) \leq d(\ell^*, \ell) + d(\ell, \mathfrak{A}[\ell]) $$
by triangle inequality.

The first term on the right hand side corresponds to how good our LM $\ell$ is, which is limited by the advancement in LM research. The second term is the gap due to our encoding algorithm and it is the focus of this study. Without knowing how large the first term is, we can still pursue a meaningful \emph{relative} imperceptibility guarantee of the form, ``it will be as hard to detect the presence of the steganographic communication as detecting the presence of $\ell$-fluent texts.'' 

We can further decompose the second term on the right hand side over each generation step and suppose $s_{< t} \coloneqq s_1\cdots s_{t-1}$ is the prefix, we can use Pinsker's inequality \cite{madhur} and additivity of Kullback–Leibler divergence (KL divergence)\footnote{We will consistently compute KL divergence in base $2$, i.e. we measure entropy in bits.} over product measures to obtain a bound via the KL divergence on each step
\begin{align*}
&d(\ell, \mathfrak{A}[\ell]) \\
& \leq \sqrt{ \frac{\ln 2}{2} \sum_{t=1}^\infty D_{KL}\left( \Prob[\cdot | s_{< t} ; \ell] \middle|\middle| \Prob\big[\cdot | s_{< t} ; \mathfrak{A}[\ell]\big] \right) } .
\end{align*}
Hence in order to obtain relative near-imperceptibility, it is sufficient to ensure that at each generation step, the effective LM $\Prob\big[\cdot | s_{< t} ; \mathfrak{A}[\ell]\big]$ is close to the base LM $\Prob[\cdot | s_{< t} ; \ell]$. (See an analogous decomposition in terms of per-step TVD in \ref{app:tvd}.)

%
\section{Analysis}
\label{sec:analysis}
Suppose $h \in \Sigma^*$ is a prefix (tokens generated up to the current step), and the base LM is $\ell$.

\subsection{\texttt{Bins}}
\label{sec:bins}
\citet{fang2017generating} divide the vocabulary into $2^k$ disjoint bins of equal sizes, $\{B_1, \cdots, B_{2^k}\}$, that is, $\Sigma = \sqcup_{i=1}^{2^k} B_i$ and $|B_i| = V / 2^k$. The partition is randomly chosen and shared between Alice and Bob. Then we split a ciphertext into $(C/k)$-many length-$k$ blocks $a_1\cdots a_{C/k}$. We encode the ciphertext by encoding each $a_i$. To encode a random block $a \sim \text{Bernoulli}(\nicefrac{1}{2})^k$, we pick a token from the $a$-th bin, i.e. $B_a$, according to $\ell$. Suppose $s$ falls in the bin $B^s$, we effectively sample a token $s$ according to
$$ \Prob[ s | h ; \texttt{Bins}[\ell] ] = \frac{1}{2^k} \frac{ \Prob[ s | h ; \ell ] }{ \Prob[ B^s | h; \ell ] } $$
and the KL divergence is
$$ D_{KL}(\Prob[ \cdot | h ; \ell ] || \Prob[ \cdot | h ; \texttt{Bins}[\ell] ]) = k - H(B) . $$
(See \ref{app:bins} for detailed derivation.)
The last term is the entropy of the partitions at the current step which is bounded between zero and $k$. Hence, the KL divergence is at most $k$ at each step. However, if the probability mass is roughly evenly distributed over each of the $2^k$ bins, then the KL divergence is close to zero. This is the \emph{implicit} assumption about fluent texts \texttt{Bins} makes.

\begin{figure}[ht]
\centering
\includegraphics[width=\columnwidth]{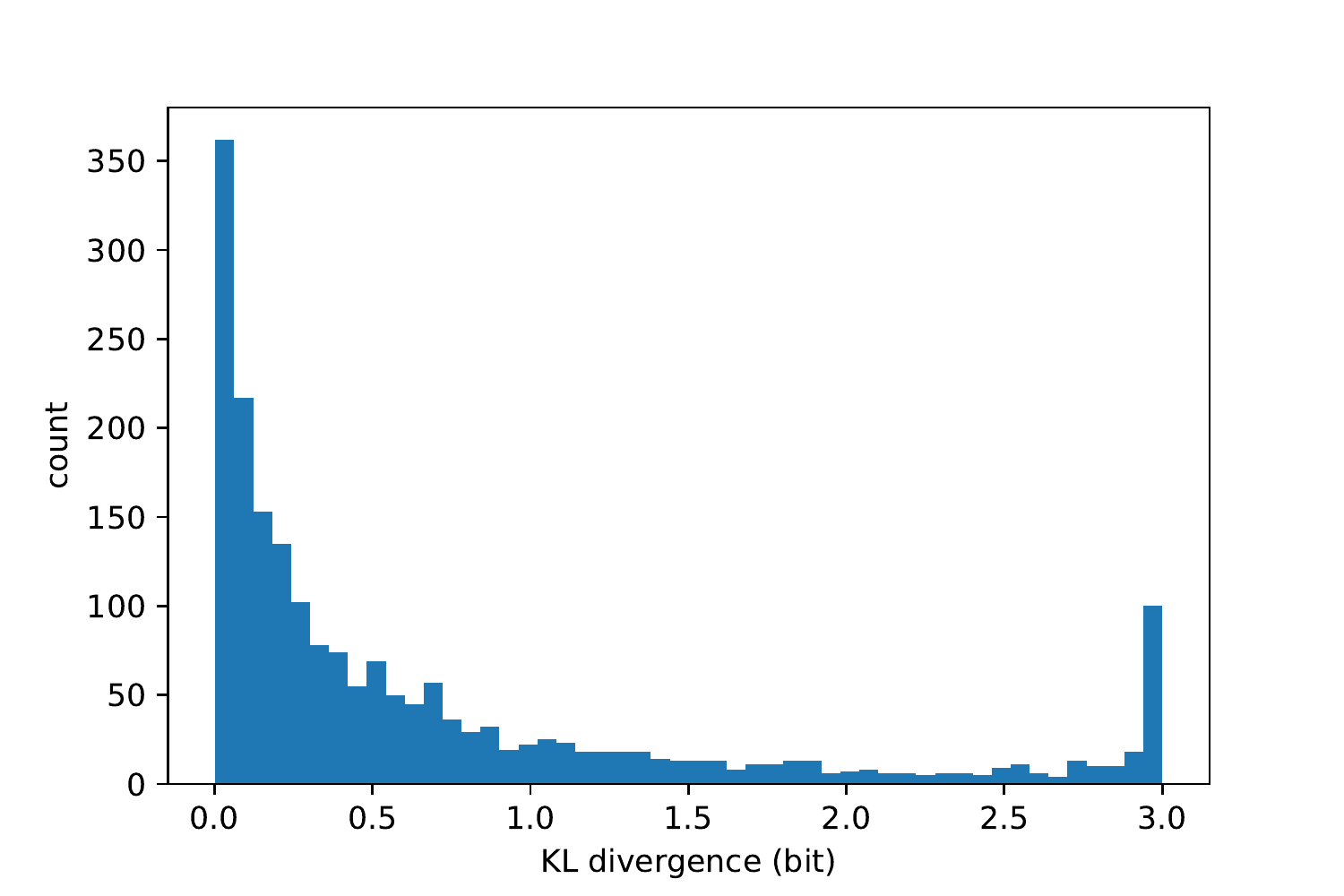}
\caption{$D_{KL}(\Prob[ \cdot | h ; \ell ] || \Prob[ \cdot | h ; \texttt{Bins}[\ell]])$ in bits over a sample of 2K tokens generated from \texttt{GPT-2-117M} with $2^3 = 8$ bins. Fewer tokens with high bits is better.}
\label{fig:kl-bins}
\end{figure}

We empirically examine how well this assumption holds. We use \texttt{GPT-2-117M} as the base LM and sample from it 50 prefixes with 40 steps each, saving 2K steps of conditional distributions. We fix a randomly chosen partition of $2^3 = 8$ bins. The computed KL divergence concentrates in the low-bit region with a second mode near 3 bit, the maximum (Fig.~\ref{fig:kl-bins}). The mean of the distribution is $0.7$ bits, meaning that in ten steps the KL bound on TVD will be vacuous, encoding about 30 bits of ciphertext.

\subsection{Variable-length coding (\texttt{VLC})}
\label{sec:vlc}
\begin{figure}[ht]
\centering
\includegraphics[width=\columnwidth]{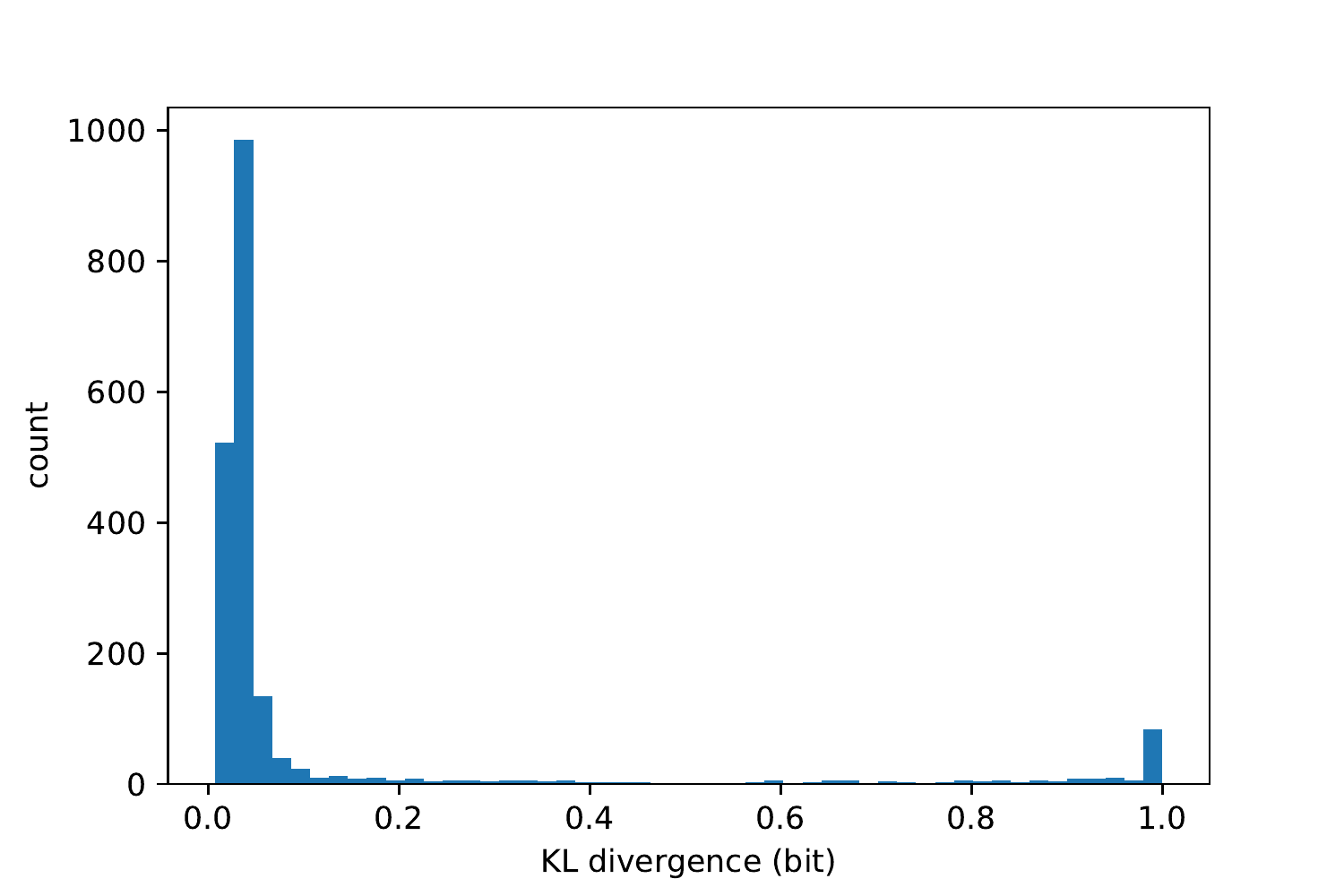}
\caption{$D_{KL}(\Prob[ \cdot | h ; \ell ] || \Prob[ \cdot | h ; \texttt{VLC}[\ell]])$ in bits over a sample of 2K tokens generated from \texttt{GPT-2-117M}. Fewer tokens with high bits is better.}
\label{fig:kl-vlc}
\end{figure}

Instead of using a fixed-length coding (one stegotext token always encodes $k$ bits in \texttt{Bins}), \texttt{VLC} encodes one or more bits per generated token \cite{yang2018rnn}. \texttt{VLC} constructs a Huffman coding of $\Sigma$ at each step according to $\Prob[ \cdot | h ; \ell ]$.\footnote{This takes $O(V \log V)$.} Then we sample a token from the constructed Huffman tree $c$ by following the bits in ciphertext starting at the root, taking the left subtree if the bit $b_i$ is zero else the right subtree until reaching a leaf. The resulting Huffman distribution $m_c$ assigns probability mass $\nicefrac{1}{2^r}$ for a token at depth $r$. Being a minimum redundancy code, the corresponding Huffman distribution has the minimum KL divergence among binary prefix-free codes \cite{huffman1952method} of at most 1 bit. But will there be steps with large KL divergence like the example ``It is on top'' in Sec.~\ref{sec:introduction}? We computed the KL divergence of Huffman codes for the same 2K samples (Fig.~\ref{fig:kl-vlc}). The mean of 0.12 bits is significantly lower than \texttt{Bins}'s but it still has a second mode near 1 bit, the maximum.

\subsection{\texttt{patient-Huffman}}
\label{sec:algorithm}
We improve \texttt{VLC} further by explicitly checking if the TVD\footnote{Computing TVD or KL divergence is $O(V)$.} (or the KL divergence) between the base LM distribution and the Huffman distribution is small enough (Algorithm~\ref{alg:patient-huffman}). If the TVD is greater than a specified threshold at the current encoding step, instead of sampling from the Huffman distribution, we sample from the base LM distribution and patiently wait for another opportunity.

\begin{algorithm}[!b]
    \caption{patient-Huffman (one encoding step)}
    \label{alg:patient-huffman}
    \begin{algorithmic}[1]
       \STATE {\bfseries Input:} a language model $\ell$, prefix $h \in \Sigma^*$, an imperceptibility threshold $\delta$, a ciphertext $b$.
       \STATE {\bfseries Return:} a stegotext from $\Sigma^*$.

       \STATE Compute the distribution of the next token $p \leftarrow \Prob[ \cdot | h ; \ell ]$.
       \STATE Construct a Huffman tree $c$ for $p$.
       \STATE Compute the TVD (or the KL divergence) between $p$ and the corresponding Huffman distribution $m_c$.
       \IF {$\text{TVD (or KL divergence)} < \delta$}
          \STATE Decode a token $w$ by consuming the ciphertext $b$ and following its bits starting at the root of Huffman tree $c$.
       \ELSE
          \STATE Sample a token $w$ according to $p$.
       \ENDIF
       \STATE Append the token to prefix $h \leftarrow h; w$
       \RETURN $h$
    \end{algorithmic}
\end{algorithm}

Clearly, this ensures that each step incurs no more additional TVD than the specified threshold $\delta$. In principle, if we set $\delta_t = o(1/t)$ for the $t$-th step, then we can bound the total TVD, \emph{guaranteeing} the relative near-imperceptibility of the generated stegotext. 

However, in practice, getting any meaningful bounds (total TVD $\ll 1$) will require setting very small $\delta_t$ and this translates to an empirical assumption that \emph{many} fluent texts' next token distributions lie \emph{arbitrarily} close to the Huffman distributions. Examining Fig.~\ref{fig:kl-vlc}, we see that there are many steps with KL divergence close to zero. This assumption, though more benign than \texttt{VLC}'s or \texttt{Bins}'s empirically, is hard to establish theoretically for fluent text.


%
\section{Discussion}
\label{sec:discussion}
We focus on the encoding algorithm in our analysis but it is not hard to see that Bob can correctly decode the ciphertext from the stegotext by running the same algorithm with the same LM and the same ciphertext block size (and other parameters if any) as Alice, e.g. \texttt{patient-Huffman} with the same threshold, and extract the unique (Huffman) code corresponding to the observed token as ciphertext.

The generic approach of embedding a ciphertext into a stegotext that has some anticipated distribution studied in this paper can very well apply to other channels such as images or audios where we can access the marginal distribution via a (deep) generative model.

Formal notions of steganographic secrecy have been studied in the cryptography community. In particular, \citet{hopper2008provably} develop a complexity theoretic notion and characterize its necessary conditions and its maximum bandwidth under a perfect sampling oracle. This is stronger than our setting where a trained LM provides us an approximate access to the marginal distribution. The information theoretic notion of imperceptibility we proposed independently is most similar to the notion of steganographic security in \cite{cachin2004information}. Further study connecting these results is needed. Of particular interest is an extension called robust steganography, where an \emph{active} adversary may alter messages, e.g. by injecting typographical errors. The stegosystems studied here are vulnerable to such attacks.

OpenAI's decision of making \texttt{GPT-2-117M} publicly available enables our empirical studies and it likely will for other studies. However, this released trained version is inferior to the full \texttt{GPT-2} model \cite{radford2019language}. While we appreciate OpenAI's general precaution and specific arguments against its release, we want to note, with this work, that its release can also offer social good by enhancing communication privacy. We advocate for the public release of strong trained models as a way to mitigate the disparity in access to both data and computational resources. 

Lastly, the full implementation of the stegosystem proposed in this work is made open-source under a permissive license.\footnote{\url{https://github.com/falcondai/lm-steganography}. We also include generated samples and illustrative examples.}

\section*{Acknowledgments}
We thank the anonymous reviewers for their suggestions. We thank David McAllester for a helpful discussion on Huffman coding. We thank Avrim Blum for bringing related works in the cryptography community to our belated attention.

\bibliography{acl2019}
\bibliographystyle{acl_natbib}

\clearpage

\appendix

\section{Appendices}
\label{sec:appendix}

\subsection{Basic facts about total variation distance}
\label{app:tvd}
Over a countable space $X$ and its discrete $\sigma$-algebra $\mathcal{P}(X)$, TVD is related to the $\ell_1$ metric, $d(p, q) = \frac{1}{2} \sum_{x \in X} |p(x) - q(x)| = \frac{1}{2}||p - q||_1$. A few useful basic facts to recall are
\begin{itemize}
    \item TVD is a metric, thus it obeys the triangle inequality.
    \item TVD is upper bounded by Kullback-Leibler (KL) divergence via Pinsker's inequality $d(p, q) \leq \sqrt{\frac{\ln 2}{2} D_{KL}(p || q) }$ where $D_{KL}(p||q) \coloneqq \Exp_x\left[ \log_2 \frac{p(x)}{q(x)} \right]$ is the KL divergence measured in bits.
    \item TVD is sub-additive over product measures $d(p_1 q_1, p_2 q_2) \leq d(p_1, p_1) + d(q_1, q_2)$, and relatedly, KL divergence is additive $ D_{KL}(p_1 q_1||p_2 q_2) = D_{KL}(p_1||p_2) + D_{KL}(q_1||q_2) $.
\end{itemize}

As an alternative to the upper bound due to KL divergence, we can also bound TVD via its sub-additivity under product measures
$$ d(\ell, \mathfrak{A}[\ell]) \leq \sum_{t=1}^\infty d\left( \Prob[\cdot | s_{< t} ; \ell], \Prob\big[\cdot | s_{< t} ; \mathfrak{A}[\ell]\big] \right) .$$ 
In fact, this can cover more general cases, such as the analogous analysis of \texttt{FLC} \cite{yang2018rnn} which zeros out everything except for the most likely tokens. We omit it due to page limit.

\subsection{GPT-2 Language Model}
\label{app:gpt-2}
The GPT-2 language model we used is a general purpose language model from OpenAI trained on WebText \cite{radford2019language}, which contains millions of web pages covering diverse topics. Citing concerns of malicious use, OpenAI only publicly released a small trained model with 117 million parameters. And that is the particular language model we use for empirical study in this work, \texttt{GPT-2-117M}. 

We choose to use GPT-2 as the base language model in our work for several reasons. First, GPT-2 is trained on a large amount of data that we do not have access to. Second, it empirically achieves state-of-the-art performance across seven challenging semantics tasks, which includes question answering, reading comprehension, summarization and translation. Third, its architecture contains many late innovations such as transformer \cite{vaswani2017attention}, instead of a recurrent neural network, and byte pair encoding for its vocabulary \cite{sennrich2016neural}.

\subsection{Derivation of Sec.~\ref{sec:bins}}
\label{app:bins}
The effective LM is equal to
\begin{align*}
  &\Prob[ s | h ; \texttt{Bins}[\ell] ] \\
  &\quad\text{By definition of \texttt{Bins}} \\
  &\quad= \Exp_a \big[\Prob[ s | s \in B_a, h ; \ell ] \big] \\
  &\quad\text{$a$ are uniformly distributed with probability $\nicefrac{1}{2^k}$} \\
  &\quad= \sum_a \frac{1}{2^k} \Prob[ s | s \in B_a, h ; \ell ] \\
  &\quad\text{The bins are disjoint, $s$ is only in $B^s$} \\
  &\quad= \frac{1}{2^k} \Prob[ s | s \in B^s, h ; \ell] \\
  &\quad\text{By definition of the marginal distribution} \\
  &\quad= \frac{1}{2^k}\frac{\Prob[ s, s \in B^s | h ; \ell ]}{\sum_{s' \in B^s} \Prob[ s', s' \in B^s | h; \ell ]} \\
  &\quad= \frac{1}{2^k} \frac{ \Prob[ s | h ; \ell ] }{ \Prob[ B^s | h; \ell ] } . \numberthis \label{eq:p-bins}
\end{align*}

The KL divergence follows as
\begin{align*}
&D_{KL}(\Prob[ \cdot | h ; \ell ] || \Prob[ \cdot | h ; \texttt{Bins}[\ell] ]) \\
&\quad = \sum_s \Prob[ s | h ; \ell ] \log_2 \frac{ \Prob[ s | h ; \ell ] } { \Prob[ s | h ; \texttt{Bins}[\ell] ] } \\
&\quad\text{Substituting in (\ref{eq:p-bins})} \\
&\quad = \sum_s \Prob[ s | h ; \ell] \log_2 \left( 2^k \, \Prob[ B^s | h ; \ell] \right) \\
&\quad = k + \sum_{s \in \Sigma} \Prob[ s | h ; \ell] \log_2 \Prob[ B^s | h ; \ell] \\
&\quad\text{$\Sigma$ is partitioned by $B = \{B_1, \cdots , B_{2^k}\}$} \\
&\quad = k + \sum_a \sum_{s \in B_a} \Prob[ s | h ; \ell] \log_2 \Prob[ B_a | h ; \ell] \\
&\quad = k + \sum_a \Prob[ B_a | h ; \ell] \log_2 \Prob[ B_a | h ; \ell] \\
&\quad = k - H(B)
\end{align*}
where $H(B)$ is the entropy of the partition $\{B_1, \cdots , B_{2^k}\}$.


\end{document}